\documentclass[a4paper]{cas-dc}

\usepackage[numbers]{natbib}
\usepackage{booktabs}
\usepackage{multirow}
\usepackage{geometry}
\usepackage{adjustbox}
\usepackage{ulem} 

\usepackage{amsmath,amsfonts}
\usepackage{algorithmic}
\usepackage{algorithm}
\usepackage{array}
\usepackage{enumitem}
\usepackage{xcolor}

\begin{document}
\let\WriteBookmarks\relax
\def\floatpagepagefraction{1}
\def\textpagefraction{.001}


\shorttitle{}    

\shortauthors{Lianshuai Guo et al.}  

\title [mode = title]{ Towards Federated Long-Tailed Graph Learning: An Energy-Guided Dual Decoupling Approach}  



%



\author[1]{Lianshuai Guo}
\ead{goluis.cc@gmail.com} 
\credit{Conceptualization, Methodology, Software, Data Curation, Investigation, Formal analysis, Writing -- original draft}

\author[1]{Zhongzheng Yuan}
\ead{genhz@mail.sdu.edu.cn} 
\credit{Software, Writing -- review \& editing}

\author[2]{Xunkai Li}
\ead{cs.xunkai.li@gmail.com} 
\credit{Methodology, Visualization, Supervision, Writing -- review \& editing}


\author[1]{Meixia Qu}
\cormark[1]
\ead{mxqu@sdu.edu.cn} 
\credit{Supervision}

\author[1]{Wenyu Wang}
\cormark[1]
\ead{hochi@sdu.edu.cn} 
\credit{Supervision}


\address[1]{Shandong University, School of Airspace Science and Engineering, Weihai 264209, China}
\address[2]{Beijing Institute of Technology, School of Computer Science and Technology, Beijing 100081, China}



\begin{abstract}
    Federated Graph Learning facilitates collaborative graph modeling across distributed clients while preserving data privacy.
    However, real-world data categories frequently exhibit long-tailed distributions. 
    Such statistical scarcity severely degrades performance in two ways: it biases the global model toward majority classes, and it structurally isolates minority nodes by submerging them in heterophilic, head-dominated neighborhoods.
    While existing methods attempt topology-agnostic statistical compensations, they often fail under data scarcity.
    Instead of recovering tail nodes, they overfit the structural noise from adjacent dominant classes, leading to representation degradation.
    To address these limitations, we propose FedEPD, a framework built on a dual decoupling paradigm that separates topological purification from semantic recalibration.
    Specifically, FedEPD utilizes distribution-aware Dirichlet energy pruning to filter spatial heterophilic edges. 
    It then overcomes Non-IID distribution shifts by extracting robust global prototypes from topologically central nodes, which are incorporated into local representations via a spatial low-pass prototype injection. 
    Furthermore, a two-stage alternating optimization strategy strictly protects majority decision boundaries while improving minority accuracy.
    Extensive experiments demonstrate that FedEPD achieves state-of-the-art performance across diverse long-tailed benchmarks, yielding absolute improvements of up to 4.97\% in Accuracy and 5.48\% in Macro-F1.
\end{abstract}







\maketitle

\section{Introduction}
\label{Introduction}
\begin{figure}[]
 \centering
 \includegraphics[width=0.4\textwidth]{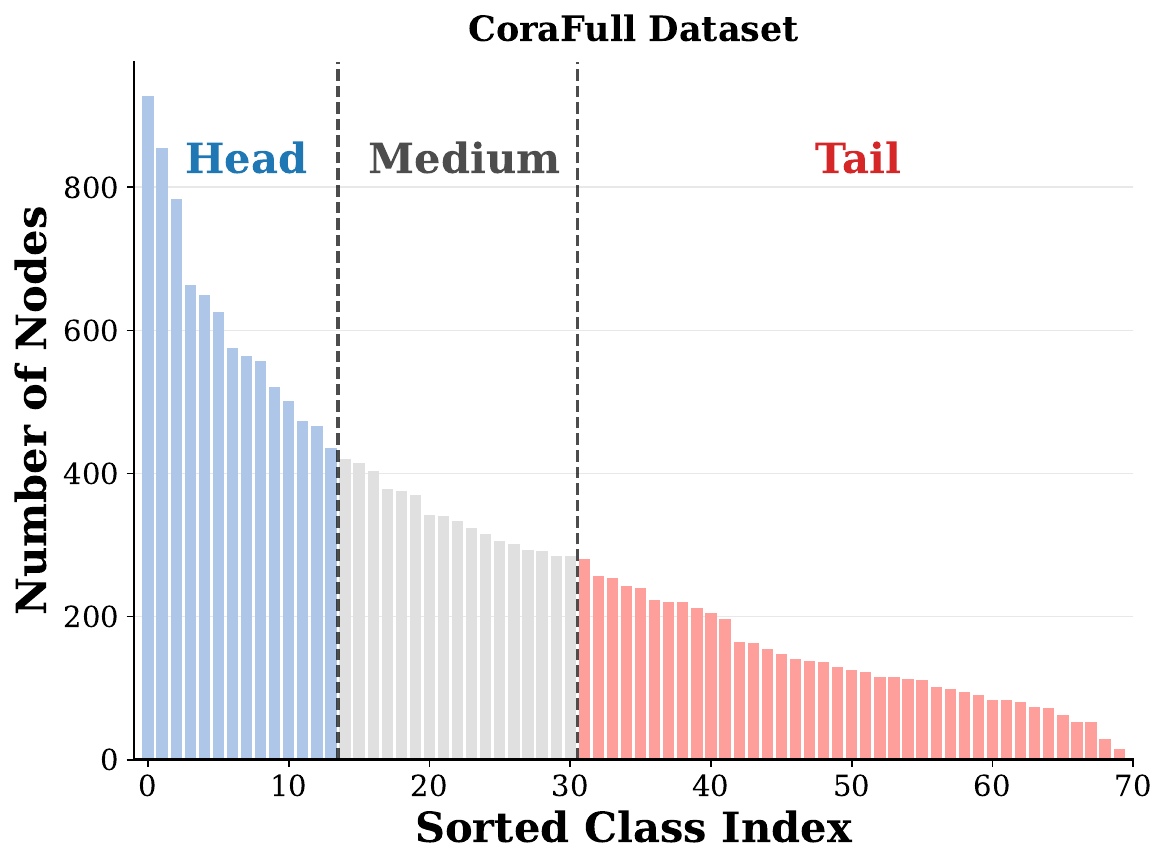}%
 \caption{Illustration of the multi-class long-tailed distribution in CoraFull, classes sorted in descending order by sample count. Node classes adhere to steep power-law dynamics. }
 \label{long-tailed}
\end{figure}

    Federated Learning establishes a distributed paradigm that enables collaborative machine learning optimization across multiple clients while preserving local data privacy \cite{li2020federated,mcmahan2017communication}. 
    Extending this decentralized framework to complex graph data, Federated Graph Learning (FGL) empowers the distributed training of Graph Neural Networks (GNNs). 
    This approach overcomes the data isolation dilemma in collaborative graph mining, allowing decentralized entities to synthesize structural information and relational dependencies without exposing raw topologies or node attributes. 
    Consequently, FGL has emerged as an indispensable technology across domains requiring data governance. 
    Recent deployments have integrated FGL into risk management in the banking sector \cite{tang2024credit}, secure molecular property prediction for collaborative drug discovery\cite{zhang2026federated,ekstrom2023accelerating}, and confidential healthcare informatics \cite{tang2024personalized}.
    However, the statistically balanced data distributions assumed in conventional FGL research diverge from empirical conditions. 
    Empirical graph networks follow power law distributions as illustrated in Fig.~\ref{long-tailed}, manifesting as multiclass long-tailed distributions where a few majority classes dominate the topology, resulting in a tail of minority classes characterized by data scarcity \cite{liu2021tail}. 
    Within decentralized Non-IID environments, this statistical imbalance is locally amplified, resulting in data sparsity where individual clients lack structural samples for specific tail categories \cite{zhang2022federated}. 
    Furthermore, this primary challenge of statistical scarcity is compounded by graph heterophily, a structural condition in empirical networks where connected nodes possess disparate semantic labels    \cite{zhu2020beyond,luan2022revisiting}. 
    Consequently, the representation learning of tail nodes is constrained by a structural conflict: these nodes are not only statistically infrequent, but their representations are also degraded by heterophilic structural noise originating from adjacent majority classes.
\begin{figure*}[t]
 \centering
 \includegraphics[width=1.0\textwidth]{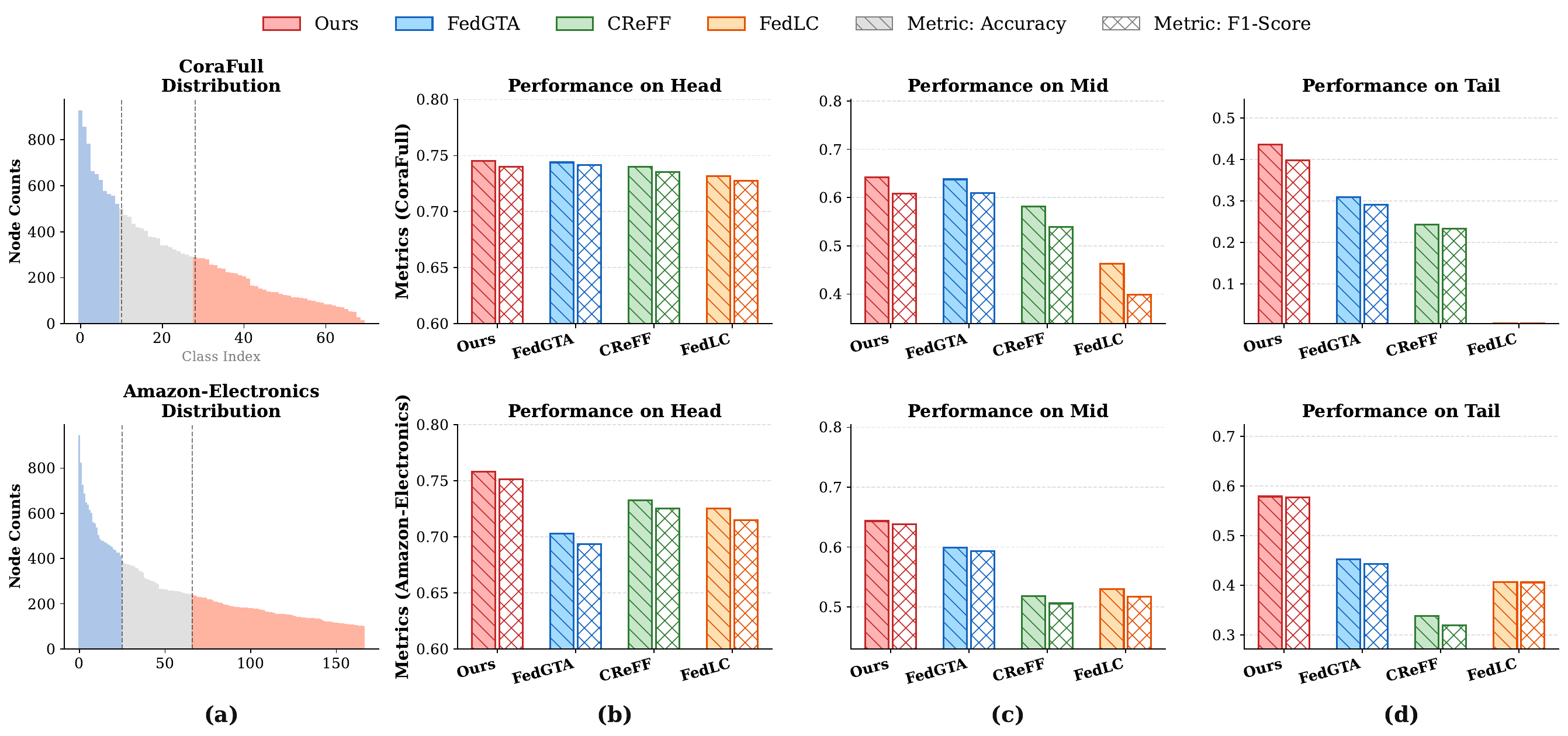}
  \vspace{-0.4cm}
 \caption{Empirical analysis of class distribution and the Majority-Minority Optimization Conflict in federated graph learning. (a) The extreme multi-class long-tailed distributions within the CoraFull and Amazon-Electronics datasets, where classes are ranked by their local node frequencies. (b)-(d) Comparative Accuracy and F1-score evaluations across the (b) Head, (c) Medium, and (d) Tail partitions for four representative federated baselines.}
 \label{empirical_result}
 \vspace{-0.4cm}
\end{figure*}
    Current methodologies designed for imbalanced learning exhibit limitations when addressing multiclass long-tailed distributions in federated graph environments \cite{zhang2025pop}. 
    A limitation of existing solutions is their tendency to treat the long-tailed challenge as statistical scarcity, overlooking the structural deficiency of minority classes. Specifically, spatial oversampling techniques attempt to synthesize tail nodes through feature interpolation \cite{zhao2021graphsmote}, while approaches operating in the feature space, such as DULL \cite{shu2025classifying}, resolve imbalance through latent feature disentanglement and partial unlearning.
    However, within isolated client subgraphs characterized by data scarcity, synthesizing or disentangling minority features without sufficient structural context generates inconsistent representations that alter the underlying topological manifold. 
    Alternatively, federated statistical calibration methods rely on logit adjustments proportional to class scarcity \cite{zhang2022federated}. Recent federated strategies like FedYoYo \cite{yan2025you} build upon this by combining autonomous knowledge distillation with distribution-adaptive logit adjustments to reduce the performance difference between centralized and federated paradigms. 
    Being independent of topology, these methods cause the classifier to expand the decision boundaries of tail categories. When these minority nodes are adjacent to majority classes, this expansion leads to overfitting of local structural noise.

    To empirically validate this representation degradation, we conduct a preliminary study on two representative long-tailed graph datasets: CoraFull and Amazon-Electronics. 
    By partitioning the nodes into Head, Medium, and Tail categories based on their class frequencies, our experimental results  as illustrated in Fig.~\ref{empirical_result} reveal a performance discrepancy: while standard models maintain high accuracy on majority classes, performance on tail categories degrades. 
    We attribute this phenomenon to the inability of existing models to reconcile statistical frequency with topological structure. 
    Specifically, we identify two limitations in current federated graph learning and propose the Federated Energy Pruning and Decoupling (FedEPD) framework to address them through a formal dual decoupling paradigm:
    
\noindent \textbf{Limitation 1.} 
    Statistical-Topological Misalignment. 
    Existing methodologies assume that long-tailed distributions can be resolved via numerical adjustments. However, in long-tailed and heterophilic environments, these topology-agnostic adjustments force the classifier to overfit structural noise, distorting decision boundaries from adjacent majority classes, leading to increased misclassification rates.

\noindent \textbf{Solution 1.} 
    Topological Purification via energy. To address this, FedEPD decouples structural denoising from semantic learning. 
    We introduce a distribution-aware Dirichlet energy pruning mechanism that filters heterophilic edges in the spatial domain. 
    This ensures that the foundational topology remains structurally denoised before representation learning occurs.

\noindent \textbf{Limitation 2.} 
    Majority-Minority Optimization Conflict. 
    While the observed empirical disparity necessitates compensation for tail categories, directly coupling these adjustments with the feature extractor forces a compromise. 
    Existing methods face an optimization dilemma: attempts to boost minority performance without architectural decoupling degrade the representations of majority classes, preventing uniform performance gains.

\noindent \textbf{Solution 2.} 
    Dual Decoupling. 
    To resolve this conflict, FedEPD executes a dual decoupling strategy across both model and representation dimensions. 
    At the model level, it employs a two-stage alternating optimization pipeline that freezes the graph encoder during calibration, protecting the foundational representations from tail-oriented logit adjustments. 
    At the representation level, it constructs robust global prototypes via a server-assisted local consensus and selectively injects them into the low-frequency semantic components of local nodes. 
    This spatial low-pass prototype injection isolates semantic compensation from high frequency structural distinctiveness, improving minority classes accuracy while preserving the decision boundaries for majority classes.

In summary, the main contributions of this paper are summarized as follows:
\begin{itemize}[leftmargin=*]
\item \textbf{New Perspective.} 
    We shift the focus of long-tailed graph learning from frequency imbalance to structural interference.
    We clarify how numerical adjustments fail in long-tailed heterophilic graphs, providing an explanation for the performance degradation on minority classes.

\item \textbf{Novel Paradigm and Framework.} 
    We propose FedEPD, a framework that implements a dual decoupling paradigm. 
    By integrating distribution-aware Dirichlet energy pruning with a two-stage alternating optimization and spatial low-pass prototype injection, our framework mitigates gradient conflicts and improves convergence stability without introducing significant computational overhead.

\item \textbf{Superior Performance.} 
    Experiments on long-tailed graph datasets confirm that FedEPD resolves the performance conflict. It preserves classification stability for head and medium classes while yielding absolute accuracy improvements of up to 11.89\% for the tail categories.
\end{itemize}

\section{Preliminary and Related Work}
\label{Preliminary}

\subsection{Notations and Problem Formulation}

    Let $\mathcal{G} = (\mathcal{V}, \mathcal{E}, \mathbf{X}, \mathbf{Y})$ denote an undirected global graph, where $\mathcal{V}$ is the set of $N$ nodes, and $\mathcal{E}$ represents the edges with adjacency matrix $\mathbf{A} \in \{0, 1\}^{N \times N}$. 
    For any node $v_i \in \mathcal{V}$, its topological neighborhood is denoted as $\mathcal{N}_i = \{v_j \in \mathcal{V} \mid (v_i, v_j) \in \mathcal{E}\}$. $\mathbf{X} \in \mathbb{R}^{N \times d}$ is the node feature matrix, and $\mathbf{Y} \in \{0, 1\}^{N \times C}$ denotes the one-hot label matrix over $C$ classes. 
    In a standard federated subgraph system with $K$ clients, the global graph is partitioned into $K$ distributed subgraphs $\{\mathcal{G}_1, \mathcal{G}_2, \dots, \mathcal{G}_K\}$. For client $k$, its local subgraph is denoted as $\mathcal{G}_k = (\mathcal{V}_k, \mathcal{E}_k, \mathbf{X}_k, \mathbf{Y}_k)$, where $n_k = |\mathcal{V}_k|$ and $\sum_{k=1}^K n_k = N$. 
    Due to privacy constraints, cross-client edges are unobservable. 
    Each client collaboratively trains a global graph neural network parameterized by $\Theta$ using only its local topology $\mathcal{E}_k$ and features $\mathbf{X}_k$.

    Within this decentralized context, we formalize the empirical long-tailed scenario where global class frequencies exhibit a power law decay \cite{yun2022lte4g}. 
    Let $N_c$ denote the total number of nodes belonging to class $c \in \{1, 2, \dots, C\}$. 
    We quantify the class imbalance via the Imbalance Ratio $\text{IR} = {\max_{c} N_c}/{\min_{c} N_c}$. 
    Assuming the classes are sorted in descending order of frequency, i.e., $N_1 \ge N_2 \ge \dots \ge N_C$, the class set is partitioned into three disjoint subsets: head classes $\mathcal{C}_{H}$, medium classes $\mathcal{C}_{M}$, and tail classes $\mathcal{C}_{T}$ \cite{liu2021tailgnn}. The empirical power law distribution satisfies $|\mathcal{C}_{H}| \ll |\mathcal{C}_{T}|$ but $\sum_{c \in \mathcal{C}_{H}} N_c \gg \sum_{c \in \mathcal{C}_{T}} N_c$. 
    This statistical scarcity induces a structural deficiency for the tail categories \cite{yun2022lte4g}. 
    For a minority node $v_i \in \mathcal{C}_T$, the lack of homophilic neighbors results in connections across different classes. Consequently, its local neighborhood $\mathcal{N}_i$ is degraded by heterophilic structural noise originating from adjacent majority classes $v_j \in \mathcal{C}_H$. This structural conflict, amplified under decentralized Non-IID settings, alters the local message passing process, motivating our dual decoupling paradigm.

\subsection{Federated Long-Tailed Learning}

    Federated Long-Tailed Learning (FedLT) extends distributed optimization to empirical scenarios characterized by statistical imbalance, where the global data distribution follows a power law \cite{chen2022towards}. 
    Unlike centralized learning, FedLT faces a compounded challenge: local data heterogeneity (Non-IID) across clients intertwines with global class imbalance. Consequently, minority classes are globally scarce and often absent on specific clients, which exacerbates gradient variance, local overfitting, and global client drift during aggregation. 
    While certain centralized approaches incorporate structural awareness to compensate for minority classes, including synthetic node and edge generation \cite{zhao2021graphsmote} and latent feature disentanglement \cite{shu2025classifying}, applying these directly to federated environments presents inherent limitations.
    Due to strict privacy constraints and the lack of a holistic data view, deploying local topological oversampling on data-scarce clients frequently amplifies heterophilic noise rather than accurately reconstructing minority structures.
    
    To mitigate this, recent works reconstruct global priors to guide local training via mechanisms such as adaptive gradient balancers \cite{xiao2023fed} or autonomous distillation \cite{zeng2023global}. Other countermeasures address the coupled Non-IID and imbalance challenges through adaptive logit adjustment \cite{lu2024fedlf} or representation learning from a decoupled perspective. By separating the optimization of the generic feature extractor from the personalized classifier, decoupled methods attempt to mitigate client drift and isolate long-tail information within the feature space \cite{zhang2026federated}. However, while these decoupled frameworks demonstrate efficacy in Euclidean domains, their topology-agnostic nature leaves the handling of structural deficiency and heterophilic noise as an open challenge in long-tailed Federated Graph Learning.

\subsection{Federated Graph Learning} 

    Federated Graph Learning (FGL) extends federated learning to graph-structured data, typically addressing two challenges: subgraph heterogeneity and missing cross-client edges. 
    To address heterogeneity, researchers have proposed various topology-aware strategies, such as personalized aggregation guided by topological similarity \cite{li2023fedgta, baek2023personalized}, decoupled knowledge distillation \cite{zhu2024fedtad}, and customized client-specific training \cite{li2024adafgl}. 
    To mitigate missing edges, other works reconstruct structural receptive fields through spatial generative mechanisms \cite{zhang2021subgraph} or secure information exchange \cite{wu2021fedgnn}. 
    Recently, attention has shifted toward the performance degradation caused by long-tailed distributions in FGL. 
    Countermeasures include adaptive gradient aggregation coupled with decoupled attention \cite{li2024fedgac} and mutual information-guided generative augmentation for minority classes \cite{li2025graphfedmig}. 

    Despite these advancements, existing frameworks exhibit limitations under the combined conditions of empirical power law distributions and topological heterophily. 
    Methods relying on numerical adjustments or traditional spatial aggregations cannot isolate semantic signals from heterophilic noise, forcing decision boundaries to overfit structural interference from adjacent majority classes. 
    Furthermore, while methods like $S^2$FGL \cite{tan2025s} explore global alignment from a spectral perspective, they are primarily designed for general federated heterogeneity and do not address class imbalance. Consequently, these approaches remain susceptible to representation degradation when majority classes dominate the topological structure. 
    This highlights a gap in the existing FGL literature: the necessity for a dual decoupling paradigm. 
    By isolating topological purification from semantic recalibration, such a framework can robustly restore tail nodes without compromising the calibrated decision boundaries of majority classes.
\begin{figure*}[t]
 \centering
 \includegraphics[width=1.0\textwidth]{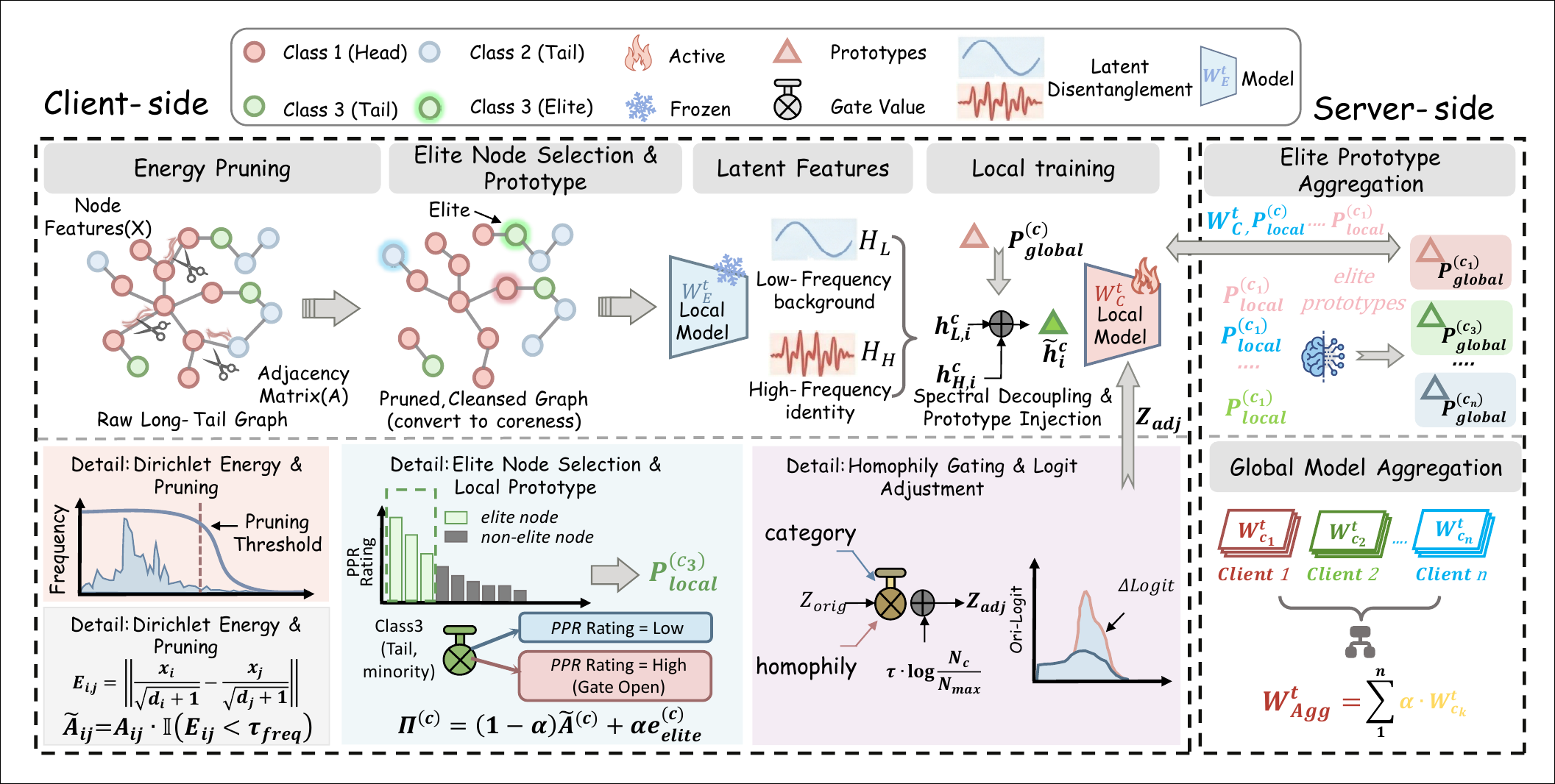}%
 \caption{The overview of our proposed FedEPD framework.}
 \label{modelFrameWork}
\end{figure*}
\section{Methodology}
\label{Methodology}
    To resolve the misalignment between statistics and topology, and the optimization conflict between majority and minority classes, we propose the FedEPD framework. 
    Unlike methods that rely on topology-agnostic numerical compensations, FedEPD reformulates the long-tailed challenge through a dual decoupling paradigm that isolates topological purification from semantic recalibration. 
    The framework operates through a two-stage pipeline. 
    In the first stage, distribution-aware Dirichlet energy pruning filters the heterophilic interference that degrades minority classes. 
    Building upon this purified topology, the second stage leverages a server-assisted local consensus to extract structurally representative nodes across the distributed system, constructing robust global prototypes that provide a reliable semantic reference. 
    Subsequently, a spatial low-pass prototype injection mechanism incorporates this global consensus strictly into the low-frequency semantic components of the representations. 
    Coupled with an alternating optimization strategy, this dual decoupling architecture protects tail nodes against heterophilic noise while preserving calibrated decision boundaries for the majority classes. 
    The overall architecture of FedEPD is illustrated in Fig.~\ref{modelFrameWork}.
\subsection{Topological Purification}
\label{sec:purification}

\noindent \textbf{Motivation.} 
    As defined in Section \ref{Preliminary}, the localized representations of tail nodes $\mathcal{C}_T$ are degraded by heterophilic message passing from adjacent majority classes $\mathcal{C}_H$. Therefore, before any semantic recalibration occurs, we must filter this heterophilic interference. 
    To avoid the computational cost of full eigendecomposition and to resist the interference of heavy-tailed outliers in graph topologies, we propose an adaptive energy-similarity pruning mechanism in the spatial domain.

\noindent \textbf{Local Energy Evaluation.} 
    Instead of operating on deep hidden representations, which may suffer from over-smoothing, we evaluate the intrinsic topological discrepancy using the raw input features. 
    For client $k$, let $\mathbf{x}_i \in \mathbb{R}^d$ denote the raw feature of node $v_i$, and $d_i$ denote its corresponding degree. 
    To prevent high-degree nodes from dominating the distance metric, we apply a degree-normalized Euclidean distance. The local Dirichlet energy ${E}_{ij}$ for any connected pair $(v_i, v_j) \in \mathcal{E}_k$ is formulated as:
\begin{equation}
{E}_{ij} = \left\| \frac{\mathbf{x}_i}{\sqrt{d_i + 1}} - \frac{\mathbf{x}_j}{\sqrt{d_j + 1}} \right\|_2^2.
\label{eq:dirichlet_energy}
\end{equation}

\noindent \textbf{Adaptive Multi-metric Fusion.} 
    Absolute energy thresholds are vulnerable to local distribution shifts. Thus, we fuse the energy perspective with semantic consistency. 
    Let ${E}_k = \{ E_{ij} \mid (v_i, v_j) \in \mathcal{E}_k \}$ denote the collection of all pairwise discrepancies within client $k$. We compute its mean $\mu_E$ and standard deviation $\sigma_E$. 
    To quantify the structural noise intensity and adaptively shift the pruning focus towards energy when noise is severe, we define a blending weight $\lambda_k$ based on the coefficient of variation $cv_E$:
\begin{equation}
cv_E = \frac{\sigma_E}{\mu_E + \epsilon}, \quad \lambda_k = \frac{cv_E^2}{1 + cv_E^2}.
\label{eq:adaptive_weight}
\end{equation}
    where $\epsilon = 10^{-12}$. Furthermore, to prevent extreme values in heavy-tailed distributions from dominating the metric, we map the disparate indicators to a uniform $[0, 1]$ scale using a non-parametric cumulative distribution. We define the standardized energy $\zeta_{ij}$ and the semantic distance $s_{ij}$ for edge $(v_i, v_j)$ as:
\begin{equation}
\zeta_{ij} = \frac{E_{ij} - \mu_E}{\sigma_E + \epsilon}, \quad s_{ij} = 1 - \frac{\mathbf{x}_i \cdot \mathbf{x}_j}{\|\mathbf{x}_i\|_2 \|\mathbf{x}_j\|_2}.
\label{eq:metric_definitions}
\end{equation}
    Let $\mathcal{Z}_k = \{\zeta_{ij} \mid (v_i, v_j) \in \mathcal{E}_k\}$ and $\mathcal{S}_k = \{s_{ij} \mid (v_i, v_j) \in \mathcal{E}_k\}$ denote the corresponding collections of these metrics within client $k$. 
    The cumulative mapping $\hat{F}_{\mathcal{X}}(x)$ for a given variable set $\mathcal{X}$ and the final fusion score $\psi_{ij}$ are formally defined as:
\begin{equation}
\begin{aligned}
    \hat{F}_{\mathcal{X}}(x) &= \frac{|\{x' \in \mathcal{X} \mid x' \le x\}|}{|\mathcal{X}|}, \\
    \psi_{ij} &= \lambda_k \cdot \hat{F}_{\mathcal{Z}_k}(\zeta_{ij}) + (1 - \lambda_k) \cdot \hat{F}_{\mathcal{S}_k}(s_{ij}).
\end{aligned}
\label{eq:fused_score}
\end{equation}
where $\mathcal{X}$ denotes an arbitrary finite collection of scalar values (e.g., $\mathcal{Z}_k$ or $\mathcal{S}_k$).

\noindent \textbf{Distribution-Aware Pruning.}
    To ensure adaptability across varying distributions, the pruning ratio $p_k$ is dynamically inferred from the structural noise intensity as $p_k = \lambda_k/2$. 
    We establish a client-specific truncation threshold $\tau_k$ at the $(1 - p_k)$-quantile of the fusion scores $\psi_{ij}$. The purified local adjacency matrix $\mathbf{A}'_k$ is formulated by filtering high-discrepancy edges across the entire local graph:
\begin{equation}
[\mathbf{A}'_k]_{ij} = \begin{cases} 
[\mathbf{A}_k]_{ij}, & \text{if } \psi_{ij} \le \tau_k \\
0, & \text{otherwise} 
\end{cases}.
\label{eq:purified_adj}
\end{equation}
    Evaluating these multi-metric scores requires $\mathcal{O}(|\mathcal{E}_k|d + |\mathcal{E}_k| \log |\mathcal{E}_k|)$ operations. 
    In our dual decoupling paradigm, this topological purification is executed once at a predefined stage transition round (specifically, the first round). 
    Consequently, the purified matrix $\mathbf{A}'_k$ remains static during subsequent optimization, ensuring that structural decoupling introduces negligible computational overhead.
\begin{algorithm}[t]
\caption{FedEPD: Federated Execution}
\label{alg:fedepd_server}
\begin{algorithmic}[1]
\REQUIRE Total communication rounds $T$, active clients $\mathcal{S}$, local graphs $\{\mathcal{G}_k\}_{k \in \mathcal{S}}$.
\ENSURE Final global model $\Theta^T$, global prototypes $\mathcal{P}^T$. 
\STATE Initialize global model weights $\Theta^0$.
\STATE Initialize global prototype set $\mathcal{P}^0 = \emptyset$.
\FOR{$t = 1, \dots, T$}
    \STATE \textit{Local Alternating Optimization \& Extraction}
    \STATE \textbf{Parallel for} client $k \in \mathcal{S}$: 
    \STATE $\Theta_k^t, \{\mathbf{u}_k^{c,t}, m_k^c\}_{c=1}^C \leftarrow \text{Client\_Execute}(k, \Theta^{t-1}, \mathcal{P}^{t-1}, t)$  
    \STATE \textit{Global Aggregation}
    \STATE Aggregate global model: $\Theta^t \leftarrow \sum_{k \in \mathcal{S}} \frac{n_k}{N} \Theta_k^t$.
    \STATE Construct global prototypes $\mathcal{P}^t = \{\mathbf{p}^c\}_{c=1}^C$ (Eq.~\eqref{eq:global_anchor}). 
    \STATE Broadcast $\Theta^t$ and $\mathcal{P}^t$ to all active clients.
\ENDFOR
\end{algorithmic}
\end{algorithm}

\begin{algorithm}[t]
\caption{FedEPD: Client Local Operations}
\label{alg:fedepd_client}
\begin{algorithmic}[1]
\REQUIRE Client $k$, global model $\Theta$, global prototypes $\mathcal{P}$, round $t$,learning rate $\eta_l$. 
\ENSURE Local model $\Theta_k$ and $\{\mathbf{u}_k^{c,t}, m_k^c\}_{c=1}^C$. 
\STATE \hrulefill
\STATE \textbf{Procedure} \textsc{Client\_Execute}($k, \Theta, \mathcal{P}, t$)  
\STATE Initialize local model $\Theta_k \leftarrow \Theta$.  
\IF{$t = T_{init}$}
    \STATE \textit{Topological Purification \& Elite Selection} 
    \STATE Calculate and cache  $\mathbf{A}'_k$ via Eq.~\eqref{eq:dirichlet_energy}--\eqref{eq:purified_adj}.
    \STATE Calculate and cache elite nodes $\mathcal{V}_{k,\text{elite}}^c$ via Eq.~\eqref{eq:iterative_ppr}. 
\ENDIF

\vspace{1mm}
\STATE \textit{Base Representation Optimization} (Eq.~\eqref{eq:loss_base}) 
\STATE Compute base loss $\mathcal{L}_{\text{base}}$ over purified graph $\mathbf{A}'_k$. 
\STATE Update full model: $\Theta_k \leftarrow \Theta_k - \eta_l \nabla_{\Theta_k} \mathcal{L}_{\text{base}}$.  

\vspace{1mm}
\IF{$\mathcal{P} \neq \emptyset$}
    \STATE \textit{Decoupled Calibration} (Eq.~\eqref{eq:normalization}--Eq.~\eqref{eq:loss_calib})  
    \STATE Compute calibration loss $\mathcal{L}_{\text{calib}}$ with frozen encoder.  
    \STATE Update classifier: $\Theta_k \leftarrow \Theta_k - \eta_l \nabla_{\Theta_{\text{classifier}}} \mathcal{L}_{\text{calib}}$.  
\ENDIF

\vspace{1mm}
\STATE \textit{Local Prototype Extraction}
\STATE Recompute $\mathbf{H}$ via updated encoder on $\mathbf{A}'_k$. 
\STATE Compute $\mathbf{u}_k^{c,t}$ over elites $\mathcal{V}_{k,\text{elite}}^c$ via Eq.~\eqref{eq:local_prototype}. 
\STATE \textbf{Return} $\Theta_k$ and $\{\mathbf{u}_k^{c,t}, m_k^c\}_{c=1}^C$.  
\end{algorithmic}
\end{algorithm}

\subsection{Server-Assisted Local Consensus}
\label{sec:consensus}

\noindent \textbf{Motivation.} 
    Adaptive spatial pruning effectively mitigates heterophilic noise; however, it simultaneously induces a structural deficiency in the local neighborhoods of tail nodes.
    Because tail nodes lack sufficient homophilic neighbors, pruning limits their receptive field. 
    To recover tail categories without altering the decision boundaries of the majority classes, a structurally denoised global semantic reference is required for each class. 
    In environments characterized by Non-IID and long-tailed distributions, individual clients do not possess a comprehensive view. 
    The server-assisted local consensus mechanism extracts features from topologically central nodes to construct a robust global prototype decoupled from local heterophilic interference.

\noindent \textbf{Local Prototype Extraction.} 
    Directly averaging node features within a local subgraph incorporates structural noise into the prototype. 
    To identify structurally representative nodes, we leverage Personalized PageRank (PPR). 
    To avoid the $\mathcal{O}(n_k^3)$ complexity of exact computation, we employ a fixed-step iterative approximation. To ensure the identified elites are semantically anchored to the target category, the random walk restarts from the local training nodes.
    For a specific class $c$ on client $k$, starting from an initial state $\boldsymbol{\pi}_k^{c(0)} = \mathbf{r}_k^c$, the scoring vector at iteration $\ell$ is computed as:
\begin{equation}
\boldsymbol{\pi}_k^{c(\ell)} =  \alpha\mathbf{r}_k^c + (1 -\alpha) \mathbf{A}_k \mathbf{D}_k^{-1} \boldsymbol{\pi}_k^{c(\ell-1)},
\label{eq:iterative_ppr}
\end{equation}
    where $\alpha = 0.15$ is the teleport probability, $\mathbf{A}_k$ is the original local adjacency matrix, $\mathbf{D}_k$ is the diagonal degree matrix, and $\mathbf{r}_k^c$ is the restart distribution uniformly initialized over the local training nodes of class $c$. Truncating this iterative approximation at a fixed maximum step $L=10$ bounds the computational complexity to $\mathcal{O}(L|\mathcal{E}_k|)$. Let $\pi_i^c$ denote the final approximated score for node $v_i$ extracted from the converged vector $\boldsymbol{\pi}_k^{c(L)}$.
    To adapt to the varying sample sizes in long-tailed distributions, the elite selection count is dynamically defined as $m_k^c = \max(1, \lfloor \rho \cdot N_k^c \rfloor)$, where $\rho \in (0, 1]$ is the elite selection ratio and $N_k^c$ is the local class size. 
    For each class $c$, client $k$ extracts the top-$m_k^c$ highest-scoring nodes to form the local elite set $\mathcal{V}_{k, elite}^c$. 

    Crucially, while the elite selection is performed on the original graph $\mathbf{A}_k$ to accurately capture global structural centrality, the actual prototype computation must be strictly decoupled from heterophilic noise. Therefore, the local prototype sum $\mathbf{u}_k^c$ is calculated using the representations encoded over the purified graph $\mathbf{A}'_k$:
\begin{equation}
\mathbf{u}_k^c = \sum_{v_i \in \mathcal{V}_{k,elite}^c} \mathbf{h}_i ,
\label{eq:local_prototype}
\end{equation}
where $\mathbf{h}_i$ represents the encoded feature of node $v_i$ obtained via the purified topology $\mathbf{A}'_k$. The sum $\mathbf{u}_k^c$ and count $m_k^c$ are transmitted to the server.

\noindent \textbf{Global Prototype Construction.} 
Upon receiving the local components, the central server aggregates them into a global prototype. 
Let $\mathcal{S}_c$ denote the subset of clients possessing class $c$. 
The global prototype $\mathbf{p}^c$ is computed as the mean of the selected elite features across the distributed system:
\begin{equation}
\mathbf{p}^c = \frac{\sum_{k \in \mathcal{S}_c} \mathbf{u}_k^c}{\sum_{k \in \mathcal{S}_c} m_k^c}.
\label{eq:global_anchor}
\end{equation}
The server broadcasts the comprehensive prototype set $\mathcal{P} = \{\mathbf{p}^1, \mathbf{p}^2, \dots, \mathbf{p}^C\}$ to all clients. These prototypes serve as semantic references, decoupled from local topological noise, for incorporation during the subsequent calibration phase.

\subsection{Decoupled Recalibration}
\label{sec:recalibration}

\noindent \textbf{Motivation.} 
    While topological purification mitigates heterophilic noise, the long-tailed distribution leaves tail classes with structural deficiency. 
    Direct feature concatenation or joint loss optimization often forces the model to compromise between majority accuracy and minority recovery. 
    Drawing on the success of decoupled long-tailed learning methods, we propose an alternating optimization strategy that disentangles base representation learning from semantic recalibration.

\noindent \textbf{Base Representation Optimization.} 
    In the first stage of local client training, the model operates under a unified encoder-classifier architecture.
    Given the purified local adjacency matrix $\mathbf{A}'_k$, the encoder generates node representations $\mathbf{H}$. The classifier then maps these representations to prediction logits $\mathbf{Z}$. 
    The model is trained using the standard cross-entropy loss:
\begin{equation}
\mathcal{L}_{base} = \frac{1}{|\mathcal{V}_k^{train}|} \sum_{i \in \mathcal{V}_k^{train}} \text{CE}(\mathbf{z}_i, y_i).
\label{eq:loss_base}
\end{equation}
\noindent \textbf{Latent Feature Calibration.} 
    To compensate for the structural deficiency of tail nodes without corrupting the decision boundaries of majority classes, we execute a decoupled calibration phase. During this phase, the graph encoder is frozen to protect the foundational representations, restricting gradient updates exclusively to the classifier.

    We introduce a Spatial Low-Pass Prototype Injection mechanism. By incorporating self-loops into the purified adjacency matrix $\mathbf{A}'_k$, the symmetric normalized adjacency matrix is directly computed as:
\begin{equation}
\hat{\mathbf{A}}'_k = (\tilde{\mathbf{D}}'_k)^{-1/2} (\mathbf{A}'_k + \mathbf{I}) (\tilde{\mathbf{D}}'_k)^{-1/2} ,
\label{eq:normalization}
\end{equation}
where $\tilde{\mathbf{D}}'_k$ is the corresponding degree matrix of $\mathbf{A}'_k + \mathbf{I}$. 

    From a graph signal processing perspective, multiplying from the left by $\hat{\mathbf{A}}'_k$ implements a spatial low-pass filter. 
    Let $\hat{\mathbf{L}}'_k = \mathbf{I} - \hat{\mathbf{A}}'_k$ denote the symmetric normalized Laplacian of the purified graph.
    Since $\hat{\mathbf{L}}'_k$ is positive semi-definite with eigenvalues $\lambda \in [0, 2]$, the spectral response of $\hat{\mathbf{A}}'_k$ is $1 - \lambda$. 
    This response attenuates high-frequency components (large $\lambda$) and preserves low-frequency signals (small $\lambda$). 
    Consequently, the operation $\mathbf{H}_{low} = \hat{\mathbf{A}}'_k \mathbf{H}$ extracts the smoothed low-frequency consensus, and its residual inherently realizes a complementary high-pass filter. 
    Naively aggregating the global prototype directly into the representations would attenuate node-specific high-frequency details. 
    To incorporate class semantics without sacrificing individual signals, we decompose the encoded representation $\mathbf{H}$ into a low-frequency consensus and a high-frequency residual:
\begin{equation}
\mathbf{H}_{low} = \hat{\mathbf{A}}'_k \mathbf{H}, \quad \mathbf{H}_{high} = \mathbf{H} - \mathbf{H}_{low}.
\label{eq:freq_decomposition}
\end{equation}

    During the training optimization, for any labeled node $v_i \in \mathcal{V}_k^{train}$, the target global prototype $\mathbf{p}^{y_i}$ is incorporated exclusively into the low-frequency component via an intensity parameter $\gamma \in (0, 1)$, while the high-frequency residual is preserved verbatim without attenuation. 
    The augmented representation $\mathbf{h}'_i$ is formulated as:
\begin{equation}
\mathbf{h}'_i = (1 - \gamma)\mathbf{h}_{i}^{low} + \gamma \mathbf{p}^{y_i} + \mathbf{h}_{i}^{high}.
\label{eq:feature_reconstruction}
\end{equation}
    Crucially, to strictly prevent label leakage, this semantic incorporation is bypassed during inference. 
    Unobserved testing nodes securely default to their original foundational representations (i.e., $\mathbf{h}'_i = \mathbf{h}_i$). 
    This design ensures that target semantic guidance is precisely applied to calibrate the classifier, while preserving the strict integrity of the evaluation protocol.

\noindent \textbf{Topology-Aware Logit Adjustment.} 
    After passing the augmented training features to the classifier, we apply a topology-aware logit adjustment to correct statistical bias. 
    To reflect structural reliability, we introduce a class homophily gate $q_k^c$. 
    For each training source node $u \in \mathcal{V}_k^{train}$, let $\mathcal{N}'_u = \{v \mid (u,v) \in \mathcal{E}_k'\}$ denote its set of valid neighbors preserved after topological purification. 
    We define its homophilic neighborhood subset as $\mathcal{N}_{u}^{+} = \{v \in \mathcal{N}'_u \mid y_v = y_u\}$. The node-level homophily $h_u$ is then estimated by the cardinality ratio:
\begin{equation}
h_u = \frac{|\mathcal{N}_{u}^{+}|}{|\mathcal{N}'_u| + \epsilon},
\label{eq:node_homophily}
\end{equation}
    where $\epsilon = 10^{-12}$. Let $\mathcal{V}_{k, c}^{train} = \{u \in \mathcal{V}_k^{train} \mid y_u = c\}$ denote the subset of local training nodes belonging to class $c$. 
    The class-level gate $q_k^c$ is computed as the mean homophily over these nodes:
\begin{equation}
q_k^c = \frac{1}{|\mathcal{V}_{k, c}^{train}|} \sum_{u \in \mathcal{V}_{k, c}^{train}} h_u .
\label{eq:class_homophily}
\end{equation}

    For classes lacking valid homophily statistics, we apply a default value of $q_k^c = 0.5$ for numerical robustness.
    Simultaneously, utilizing the local class count $N_k^c = |\mathcal{V}_{k, c}^{train}|$ and the maximum class count $N_k^{\max}$, we combine the homophily gate $q_k^c$ and a tunable scaling intensity $\mu$ to derive the topology-aware shift margin $\Delta_k^c$ and the calibrated logit scalar $\tilde{z}_{i,c}$ for class $c$ in a unified step:
\begin{equation}
\Delta_k^c = \mu \cdot q_k^c \cdot \log \left( \frac{N_k^{\max} + \epsilon}{N_k^c + \epsilon} \right), \quad \tilde{z}_{i,c} = z_{i,c} + \Delta_k^c .
\label{eq:logit_adjustment}
\end{equation}
    The calibration phase is optimized by updating only the classifier weights based on the adjusted full logit vector $\tilde{\mathbf{z}}_i$:
\begin{equation}
\mathcal{L}_{calib} = \frac{1}{|\mathcal{V}_k^{train}|} \sum_{i \in \mathcal{V}_k^{train}} \text{CE}(\tilde{\mathbf{z}}_i, y_i).
\label{eq:loss_calib}
\end{equation}
    By decoupling these objectives, FedEPD ensures that the feature extractor captures robust topological patterns via $\mathcal{L}_{base}$, while the classifier maintains calibrated decision boundaries via $\mathcal{L}_{calib}$ without corrupting the representation backbone.
    The complete algorithmic workflow is presented in Algorithm \ref{alg:fedepd_server} and Algorithm \ref{alg:fedepd_client}.
\begin{table*}[t]
\begin{center}
    \centering
    \renewcommand{\arraystretch}{1.15}
    \caption{The statistical information of the experimental datasets.}
    \label{datasetSplit}
    \resizebox{\textwidth}{!}{
    \begin{tabular}{c cccccc c}
    \hline
        \textbf{Dataset}   &\textbf{Nodes}      & \textbf{Features}  & \textbf{Edges}     & \textbf{Classes} & \textbf{IR}      & \textbf{Train/Val/Test}    & \textbf{Description} \\ \hline
        CoraFull           & 19,793     & 8,710     & 126,842   & 70      & 61.87   & 60\%/20\%/20\%      & citation network \\ 
        ogbn-arxiv         & 169,343    & 128       & 1,166,243 & 40      & 942.10  & 60\%/20\%/20\%     & citation network \\ \hline
        Amazon-Electronics & 42,318     & 8,669     & 129,430   & 167     & 9.36    & 60\%/20\%/20\%       & copurchase graph \\ 
        Amazon-Clothing    & 24,919     & 9,034     & 208,279   & 77      & 10.29   & 60\%/20\%/20\%      & copurchase graph \\ \hline
        Roman-Empire       & 22,662     & 300       & 65,854    & 18      & 10.14   & 60\%/20\%/20\%      & heterophilic graph \\ \hline
        Email              & 1,005      & 128       & 50,481    & 42      & 107.00  & 60\%/20\%/20\%     & communication network \\ \hline
    \end{tabular}}
\end{center}
\vspace{-2pt}
\end{table*}
\begin{table*}[t]
\centering
\renewcommand{\arraystretch}{1.2}
\caption{Performance comparison on six datasets. Results are reported in percentage (\%) as the mean and standard deviation over five independent runs. \textbf{Bold} indicates the best performance among federated methods, and underline denotes the second-best.}
\label{main_results_categorized}

\begin{adjustbox}{width=\textwidth}
\begin{tabular}{c|ccc|ccc|ccc}
\hline
\multirow{2}{*}{\textbf{Method}} & \multicolumn{3}{c|}{\textbf{CoraFull}} & \multicolumn{3}{c|}{\textbf{Amazon-Electronics}} & \multicolumn{3}{c}{\textbf{Amazon-Clothing}} \\
\cline{2-10}
& Acc & bAcc & Macro-F1 & Acc & bAcc & Macro-F1 & Acc & bAcc & Macro-F1 \\
\hline
\hline
FedAvg      & 65.64$\pm$0.1 & 56.26$\pm$0.2 & 55.00$\pm$0.3 & 62.33$\pm$0.2 & 59.30$\pm$0.2 & 59.15$\pm$0.3 & 68.37$\pm$0.3 & 65.77$\pm$0.2 & 66.05$\pm$0.2 \\
FedProto    & 51.02$\pm$0.7 & 39.26$\pm$0.2 & 39.07$\pm$0.6 & 39.14$\pm$1.0 & 32.89$\pm$1.1 & 32.66$\pm$1.1 & 58.33$\pm$0.4 & 54.17$\pm$0.5 & 53.61$\pm$0.5 \\
FedGTA      & \underline{66.09}$\pm$0.1 & \underline{58.80}$\pm$0.2 & \underline{60.05}$\pm$0.1 & 62.89$\pm$0.1 & 59.92$\pm$0.0 & 60.07$\pm$0.1 & \underline{70.74}$\pm$0.2 & \underline{68.63}$\pm$0.2 & \underline{68.34}$\pm$0.2 \\
\hline
GraphSMOTE  & 64.09$\pm$0.2 & 53.66$\pm$1.0 & 53.13$\pm$1.2 & \underline{64.04}$\pm$0.2 & \underline{61.37}$\pm$0.1 & \underline{60.45}$\pm$0.1 & 65.06$\pm$0.5 & 58.80$\pm$0.8 & 58.51$\pm$1.3 \\
HieTail     & 65.86$\pm$0.3 & 55.53$\pm$0.4 & 56.66$\pm$0.4 & 61.00$\pm$0.4 & 56.48$\pm$0.4 & 56.20$\pm$0.4 & 68.30$\pm$0.6 & 63.03$\pm$1.1 & 62.93$\pm$1.1 \\
LWS         & 65.54$\pm$0.3 & 54.99$\pm$0.5 & 56.98$\pm$0.5 & 63.83$\pm$0.5 & 59.89$\pm$0.8 & 59.42$\pm$0.3 & 70.42$\pm$0.4 & 65.59$\pm$0.6 & 66.08$\pm$0.4 \\
\hline
FedLC       & 57.85$\pm$0.1 & 42.58$\pm$0.3 & 41.73$\pm$0.2 & 58.37$\pm$0.1 & 52.39$\pm$0.1 & 52.57$\pm$0.2 & 52.01$\pm$0.2 & 44.56$\pm$4.8 & 43.91$\pm$5.8 \\
FedGAC      & 57.59$\pm$0.5 & 54.36$\pm$0.7 & 52.04$\pm$0.9 & 57.05$\pm$0.4 & 57.34$\pm$0.4 & 54.31$\pm$0.4 & 57.64$\pm$0.9 & 59.09$\pm$1.2 & 56.38$\pm$0.7 \\
CReFF       & 63.86$\pm$0.2 & 54.38$\pm$0.2 & 56.01$\pm$0.2 & 58.67$\pm$0.3 & 52.77$\pm$0.5 & 53.30$\pm$0.5 & 70.14$\pm$0.1 & 66.20$\pm$0.1 & 66.59$\pm$0.3 \\
FedSpray    & 64.14$\pm$0.8 & 53.40$\pm$0.3 & 54.53$\pm$0.6 & 61.65$\pm$0.4 & 56.25$\pm$0.5 & 56.32$\pm$0.7 & 69.04$\pm$0.2 & 65.13$\pm$0.3 & 65.25$\pm$0.2 \\
GraphFedMig & 53.29$\pm$1.2 & 42.82$\pm$0.4 & 45.46$\pm$0.2 & 47.71$\pm$0.5 & 40.50$\pm$0.9 & 41.95$\pm$0.8 & 63.36$\pm$0.1 & 60.91$\pm$0.1 & 60.08$\pm$0.3 \\
\hline
\hline
\textbf{Ours} & \textbf{66.71$\pm$0.1} & \textbf{60.00$\pm$0.2} & \textbf{61.11$\pm$0.2} & \textbf{67.96$\pm$0.1} & \textbf{63.99$\pm$0.1} & \textbf{65.09$\pm$0.2} & \textbf{72.79$\pm$0.2} & \textbf{71.11$\pm$0.3} & \textbf{71.70$\pm$0.3} \\
\hline
Improve & $\color{red}{\Uparrow}$ 0.62\% & $\color{red}{\Uparrow}$ 1.20\% & $\color{red}{\Uparrow}$ 1.06\% & $\color{red}{\Uparrow}$ 3.92\% & $\color{red}{\Uparrow}$ 2.62\% & $\color{red}{\Uparrow}$ 4.64\% & $\color{red}{\Uparrow}$ 2.05\% & $\color{red}{\Uparrow}$ 2.48\% & $\color{red}{\Uparrow}$ 3.36\% \\
\hline
\end{tabular}
\end{adjustbox}

\vspace{0.4cm}

\begin{adjustbox}{width=\textwidth}
\begin{tabular}{c|ccc|ccc|ccc}
\hline
\multirow{2}{*}{\textbf{Method}} & \multicolumn{3}{c|}{\textbf{Roman-Empire}} & \multicolumn{3}{c|}{\textbf{ogbn-arxiv}} & \multicolumn{3}{c}{\textbf{Email}} \\
\cline{2-10}
& Acc & bAcc & Macro-F1 & Acc & bAcc & Macro-F1 & Acc & bAcc & Macro-F1 \\
\hline
\hline
FedAvg      & 45.03$\pm$0.2 & 36.74$\pm$0.2 & 36.00$\pm$0.2 & 61.63$\pm$0.2 & 31.64$\pm$0.4 & 31.88$\pm$0.5 & 36.93$\pm$1.0 & 20.56$\pm$0.6 & 18.00$\pm$0.8 \\
FedProto    & 26.35$\pm$0.2 & 16.20$\pm$0.7 & 15.00$\pm$1.0 & 56.30$\pm$0.9 & 24.09$\pm$0.7 & 23.23$\pm$0.6 & 41.80$\pm$0.2 & 26.78$\pm$0.2 & 27.72$\pm$0.3 \\
FedGTA      & 45.72$\pm$0.2 & 37.36$\pm$0.2 & 36.57$\pm$0.3 & \underline{63.96}$\pm$0.1 & 34.42$\pm$0.1 & 34.52$\pm$0.1 & \underline{47.05}$\pm$0.8 & 30.18$\pm$0.5 & 29.19$\pm$0.7 \\
\hline
GraphSMOTE  & 42.61$\pm$0.5 & 36.08$\pm$0.3 & 33.21$\pm$0.3 & 61.49$\pm$0.1 & \underline{42.52}$\pm$0.2 & \underline{40.69}$\pm$0.2 & 22.76$\pm$1.7 & 13.09$\pm$2.1 & 9.40$\pm$1.9 \\
HieTail     & 43.27$\pm$0.3 & 34.41$\pm$0.5 & 33.68$\pm$0.4 & 59.12$\pm$0.3 & 25.22$\pm$0.3 & 24.50$\pm$0.2 & 38.44$\pm$0.7 & 21.10$\pm$0.2 & 18.42$\pm$0.4 \\
LWS & 42.29$\pm$0.4 & 33.95$\pm$1.1 & 32.50$\pm$0.8 & 53.99$\pm$0.1 & 20.63$\pm$0.2 & 19.00$\pm$0.4 & 24.27$\pm$2.9 & 11.07$\pm$2.1 & 8.00$\pm$2.7 \\
\hline
FedLC       & 41.29$\pm$1.0 & 26.06$\pm$0.9 & 25.89$\pm$0.9 & 52.94$\pm$0.2 & 20.26$\pm$0.4 & 18.47$\pm$0.3 & 37.51$\pm$1.6 & 21.09$\pm$2.2 & 18.57$\pm$2.7 \\
FedGAC      & \underline{48.58}$\pm$0.8 & \underline{42.64}$\pm$0.2 & \textbf{40.16$\pm$0.2} & OOM & OOM & OOM & 24.74$\pm$0.5 & 13.05$\pm$2.1 & 9.21$\pm$1.4 \\
CReFF       & 38.99$\pm$0.3 & 29.56$\pm$1.3 & 28.09$\pm$1.6 & 51.74$\pm$0.3 & 19.28$\pm$0.1 & 17.63$\pm$0.2 & 36.00$\pm$0.9 & 20.59$\pm$1.4 & 18.41$\pm$1.5 \\
FedSpray    & 41.29$\pm$0.7 & 32.98$\pm$0.9 & 31.35$\pm$1.1 & 58.14$\pm$0.2 & 23.77$\pm$0.3 & 23.22$\pm$0.6 & 37.40$\pm$0.7 & 22.93$\pm$1.4 & 20.54$\pm$1.3 \\
GraphFedMig & 29.90$\pm$0.3 & 22.90$\pm$0.6 & 22.69$\pm$1.0 & 43.37$\pm$0.8 & 15.88$\pm$0.6 & 16.97$\pm$1.2 & 45.99$\pm$0.3 & \underline{30.30}$\pm$0.2 & \underline{30.12}$\pm$0.5 \\
\hline
\hline
\textbf{Ours} & \textbf{49.44$\pm$0.1} & \textbf{43.74$\pm$0.2} & \underline{39.21}$\pm$0.2 & \textbf{68.93$\pm$0.1} & \textbf{43.57$\pm$0.1} & \textbf{46.17$\pm$0.1} & \textbf{48.08$\pm$0.6} & \textbf{31.05$\pm$1.0} & \textbf{31.16$\pm$1.4} \\
\hline
Improve & $\color{red}{\Uparrow}$ 0.86\% & $\color{red}{\Uparrow}$ 1.10\% & $\Downarrow$ 0.95\% & $\color{red}{\Uparrow}$ 4.97\% & $\color{red}{\Uparrow}$ 1.05\% & $\color{red}{\Uparrow}$ 5.48\% & $\color{red}{\Uparrow}$ 1.03\% & $\color{red}{\Uparrow}$ 0.75\% & $\color{red}{\Uparrow}$ 1.04\% \\
\hline
\end{tabular}
\end{adjustbox}
\end{table*}
\section{Experiments}\label{Experiments}
    To evaluate the effectiveness and robustness of the FedEPD framework, all empirical evaluations are implemented within the OpenFGL~\cite{li2024openfgl} framework. 
    These experiments address the following research questions: 
    \textbf{Q1:} Does FedEPD demonstrate superiority across various graph topologies compared to existing baselines? 
    \textbf{Q2:} Under naturally long-tailed distributions, how does FedEPD restore tail classes without sacrificing majority class performance? 
    \textbf{Q3:} How does each module contribute to the performance of FedEPD? 
    \textbf{Q4:} How do varying hyperparameter settings impact the performance of FedEPD? 
    \textbf{Q5:} How efficient is the decoupled training strategy of FedEPD in terms of communication cost and convergence?
\subsection{Experimental Setup}
\noindent\textbf{Datasets.} 
    Unlike previous works that inject class imbalance via random downsampling, we evaluate our framework in empirical settings using datasets with naturally severe long-tailed properties. 
    We select six datasets covering diverse topological characteristics and scales. 
    These encompass two citation networks, CoraFull~\cite{bojchevski2017deep} and ogbn-arxiv~\cite{hu2020open}; 
    two product purchasing networks, Amazon-Clothing~\cite{mcauley2015inferring} and Amazon-Electronics~\cite{mcauley2015inferring}; 
    communication graph Email~\cite{yin2017local}; 
    and heterophilic Roman-Empire graph~\cite{platonov2023critical}. 
    The natural class imbalance challenges the models' ability to capture minority semantics without overfitting majority structural noise. 
    Detailed characteristics of the six datasets are reported in Table~\ref{datasetSplit}, where the empirical long-tailed distribution is quantified by the global Imbalance Ratio (IR). 
    The IR is defined as the ratio of sample sizes between the most frequent majority class and the rarest minority class in the unpartitioned graph:
\begin{equation}
    \mathrm{IR} = \frac{\max_{c \in \mathcal{C}} N_c}{\min_{c \in \mathcal{C}} N_c},
\label{eq:imbalance_ratio}
\end{equation}
where $N_c$ denotes the total number of nodes belonging to class $c$.

\noindent\textbf{Baselines.} 
    We compare FedEPD against eleven baselines, categorized into three paradigms: 
    Standard FL / FGL: This category evaluates standard performance without specific long-tailed compensation, including FedAvg \cite{mcmahan2017communication}, FedProto \cite{tan2022fedproto}, and FedGTA \cite{li2023fedgta}. 
    Centralized Long-Tailed Graph Learning: This category represents methods designed for centralized graph imbalance, including GraphSMOTE \cite{zhao2021graphsmote}, HierTail \cite{wang2024mastering}, and Learnable Weight Scaling (LWS) \cite{kang2019decoupling}. To adapt these methods to the federated environment, we integrate them with the FedAvg aggregation protocol. 
    Federated Long-Tailed Graph Learning: This category encompasses baselines addressing both Non-IID and class imbalance. It includes methods designed for Long-Tailed Federated Graph Learning, namely FedGAC \cite{li2024fedgac} and GraphFedMig \cite{li2025graphfedmig}, alongside federated long-tailed frameworks including FedLC \cite{zhang2022federated}, CreFF \cite{shang2022federated}, and FedSpray \cite{fu2024federated}. 
    
\noindent\textbf{Metrics.} 
    We employ Overall Accuracy (Acc), Macro-F1 score, and Balanced Accuracy (bAcc) as evaluation metrics. 
    Because Macro-F1 and bAcc assign equal weight to each class regardless of sample size, they prevent majority class domination and provide an objective evaluation of classification performance under long-tailed distributions.
    
\noindent\textbf{Experimental Settings.} 
    For baseline methods, we follow the hyperparameters recommended by the original authors. 
    Across all experiments, we employ a two-layer GCN backbone with a hidden dimension of 64 and a learning rate of $1e-2$. The elite selection ratio $\rho$ for local prototype extraction is set to $0.1$, ensuring that prototypes are constructed from the most structurally central $10\%$ of nodes for each class. The framework is trained for 200 communication rounds, with each client performing 3 local epochs per round. To simulate federated data silos, the training nodes of the graphs are distributed across 10 clients. 
    
\noindent\textbf{Experiment Environment.} 
    All evaluations are conducted on a workstation equipped with an Intel Core i7-13700K CPU, an NVIDIA GeForce RTX 3090 GPU (24 GB VRAM), and 64 GB memory. The software environment is built on Ubuntu 22.04 LTS, running CUDA 12.6.
\subsection{Overall Performance}
    To answer \textbf{Q1}, we evaluate our proposed framework against eleven baselines across six datasets. 
    As detailed in Table \ref{main_results_categorized}, the graphs are partitioned into 10 clients using the Louvain algorithm, which induces structural and Non-IID data distributions by grouping communities. 
    Given the skewed nature of the data, Overall Accuracy can be dominated by majority classes. 
    Therefore, we primarily focus on Balanced Accuracy (bAcc) and the Macro-F1 score to evaluate semantic recovery alongside overall decision boundaries.

    The empirical results demonstrate the overarching superiority of FedEPD across the evaluated benchmarks. 
    Specifically, our method outperforms all baselines across the six datasets on Balanced Accuracy (bAcc), confirming its effectiveness in unbiased semantic recovery. 
    Even on the highly heterophilic Roman-Empire dataset, where FedGAC marginally leads the Macro-F1 score (40.16\% vs. 39.21\%), FedEPD achieves the best overall accuracy and balanced accuracy, with comparable Macro-F1 performance.
    Furthermore, on the Amazon-Electronics dataset, our method achieves a Macro-F1 of 65.09\% and a bAcc of 63.99\%, yielding an absolute improvement of 5.02\% and 4.07\% respectively over the best-performing standard federated graph baseline, FedGTA. 

    In standard graph learning paradigms, GNNs aggregate features over the topology, making them susceptible to structural noise propagation in long-tailed scenarios. 
    Because tail nodes lack sufficient homophilic neighbors, their representations are degraded by the influx of messages from majority classes. This vulnerability is evident in federated adaptations of centralized long-tailed methods, such as GraphSMOTE and LWS. 
    On the heterophilic Roman-Empire dataset, these methods, which rely on spatial interpolation or synthetic neighbor generation, exhibit performance degradation, dropping to between 32\% and 34\% in Macro-F1.

    By utilizing Dirichlet energy to truncate heterophilic connections, FedEPD constructs purified local topologies. Subsequently, through spatial low-pass prototype injection and two-stage alternating optimization, it incorporates robust global prototypes into the low-frequency semantics.
    This dual decoupling strategy prevents representation degradation and mitigates heterophilic interference. 
    Finally, on large-scale graphs such as ogbn-arxiv, our method yields a 5.48\% absolute F1 gain over GraphSMOTE, demonstrating that scalable federated graph learning can effectively address severe long-tailed classification challenges.
\begin{figure}[]
    \centering
     \includegraphics[width=0.45\textwidth]{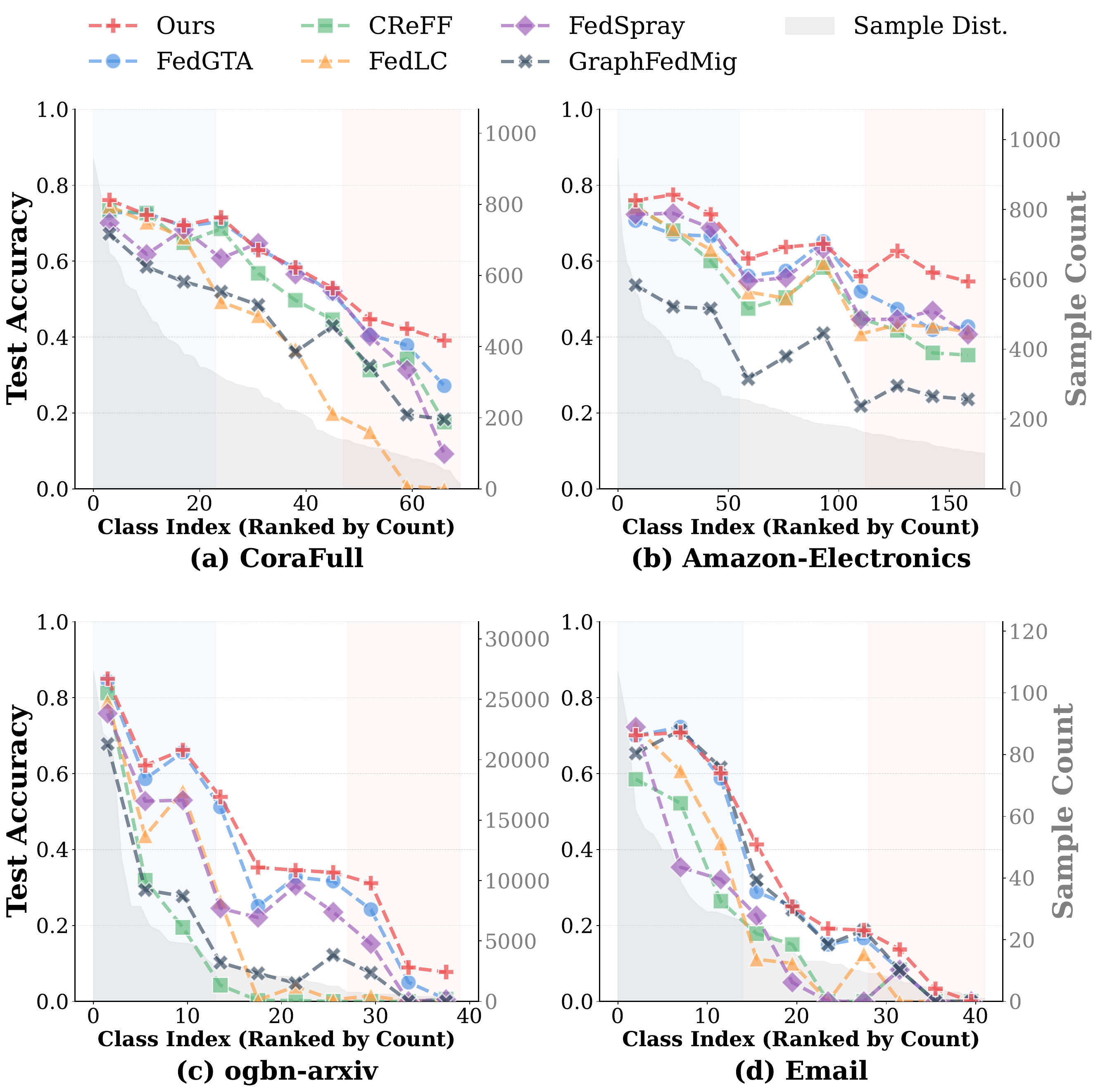}%
    \caption{Category-level test accuracy across the long-tailed distribution on four datasets. The background gray shaded area illustrates the class sample sizes (right y-axis) sorted in descending order. To clearly visualize the performance trends, the sorted classes are uniformly aggregated into 10 bins along the x-axis. Solid lines track the average test accuracy (left y-axis) of different methods within each bin. The colored background spans visually delineate the Head, Medium, and Tail regions.}
    \label{perclass_performance}
    \vspace{-10pt}
\end{figure}
\subsection{Head/Tail Performance}
    To answer \textbf{Q2}, we conduct a detailed evaluation of category-level accuracy across the frequency spectrum. 
    As illustrated in Fig.~\ref{perclass_performance}, we visualize this dynamic using a line chart superimposed on a descending background bar chart representing class sample sizes. 
    This visualization allows us to observe performance trajectories from dominant head categories to extreme tail segments. 

    For instance, on the CoraFull dataset, while methods like FedGTA and CReFF maintain high accuracy on the head classes, their performance declines to 27.16\% and 17.58\%, respectively, in the extreme tail region, with FedLC dropping to near zero. 
    In contrast, FedEPD mitigates this decline, improving the aggregated tail performance (the average of the last three bins) to 41.99\% while sustaining a competitive head performance of 72.58\%. 
    This advantage is also evident on the Amazon-Electronics dataset, where FedEPD scores an average of 75.26\%, 61.24\%, and 58.09\% across the head, medium, and tail segments. Notably, on the most infrequent class, FedEPD achieves 54.66\%, outperforming FedGTA (42.77\%) and GraphFedMig (23.58\%).
    This granular analysis is further corroborated by observations on the large-scale ogbn-arxiv dataset and the heterophilic Email network. On ogbn-arxiv, existing baselines exhibit representation degradation in the tail; methods like CReFF and FedLC degrade to 0.0\% accuracy on certain extreme tail categories. Conversely, FedEPD maintains an accuracy of 31.14\% and 8.91\% in the highly imbalanced eighth and ninth categories, respectively, where baseline predictions fail to generalize. Across all evaluated graphs, the visualization demonstrates a consistent outcome: our framework yields performance improvements for tail categories without sacrificing majority stability.
\begin{table*}[t]
\centering
\renewcommand{\arraystretch}{1.2}
\caption{Ablation study of the proposed FedEPD framework. Results are reported in percentage (\%) as the mean and standard deviation. \textbf{Bold} indicates the best performance.}
\label{tab:ablation}
\begin{adjustbox}{width=\textwidth}
\begin{tabular}{c|ccc|ccc|ccc}
\hline
\multirow{2}{*}{\textbf{Method}} & \multicolumn{3}{c|}{\textbf{CoraFull}} & \multicolumn{3}{c|}{\textbf{Amazon-Electronics}} & \multicolumn{3}{c}{\textbf{Roman-Empire}} \\
\cline{2-10}
& Acc & bAcc & M-F1 & Acc & bAcc & M-F1 & Acc & bAcc & M-F1 \\
\hline
\hline
w/o Consensus & 66.31$\pm$0.2   & 59.98$\pm$0.2 & 61.03$\pm$0.3 & 67.73$\pm$0.1 & 63.73$\pm$0.1 & 64.92$\pm$0.2 & 47.40$\pm$0.2 & 37.26$\pm$0.2 & 36.75$\pm$0.0 \\
w/o Topological Purification    & 65.87$\pm$0.1 & 59.26$\pm$0.6 & 59.19$\pm$0.6 & 60.62$\pm$0.0 & 57.69$\pm$0.1 & 58.18$\pm$0.1 & 33.18$\pm$0.3 & 26.21$\pm$0.5 & 26.09$\pm$0.5 \\
w/o Decoupled Recalibration     & 65.77$\pm$0.2 & 58.98$\pm$0.1 & 61.15$\pm$0.1 & 67.04$\pm$0.1 & 63.01$\pm$0.2 & 64.27$\pm$0.2 & 48.37$\pm$0.2 & 38.74$\pm$0.1 & 36.94$\pm$0.2 \\
\hline
\textbf{FedEPD} & \textbf{66.73$\pm$0.2} & \textbf{60.03$\pm$0.1} & \textbf{61.19$\pm$0.1} & \textbf{68.05$\pm$0.1} & \textbf{64.07$\pm$0.2} & \textbf{65.25$\pm$0.1} & \textbf{49.34$\pm$0.0} & \textbf{43.35$\pm$0.2} & \textbf{39.16$\pm$0.2} \\
\hline
\end{tabular}
\end{adjustbox}
\end{table*}
\subsection{Ablation Study}
    To answer \textbf{Q3}, we conduct an ablation study on the CoraFull, Amazon-Electronics, and Roman-Empire datasets, as detailed in Table~\ref{tab:ablation}. 
    By systematically removing core components, we design three specific variants: w/o Consensus: removing the server-assisted local consensus (including prototype injection), thereby depriving tail nodes of global semantic references; w/o Topological Purification: omitting the Dirichlet energy purification, leaving the raw heterophilic topology intact; and w/o Decoupled Recalibration: replacing the two-stage decoupled calibration (logit adjustment) with a standard federated optimization. This methodology isolates the performance impact of each mechanism. The results reveal performance degradation across all variants, confirming the necessity of each module.

    Notably, eliminating the Dirichlet energy purification (w/o Topological Purification) forces the model to absorb heterophilic structural noise, leading to the most significant accuracy reduction. This is pronounced on the heterophilic Roman-Empire dataset, where the Macro-F1 score decreases from 39.16\% to 26.09\%, and bAcc drops from 43.35\% to 26.21\%. A similar decline is observed on the Amazon-Electronics dataset, where Macro-F1 falls from 65.25\% to 58.18\%. This degradation validates that filtering structural interference prior to representation learning is a prerequisite for robust classification.

    Furthermore, the removal of the consensus module (w/o Consensus) triggers a decline across all metrics. This highlights the role of constructing a robust global prototype to provide semantic reference for tail nodes within Non-IID environments. Finally, the performance reduction in the w/o Decoupled Recalibration variant confirms that joint optimization forces the feature encoder to compromise between topological message passing and semantic classification. This coupling leads to representation distortion and the reemergence of the performance conflict between majority and minority classes. Ultimately, the superior performance of the full FedEPD framework demonstrates that these modules are complementary, and the absence of any single component compromises the classification capability.
\begin{figure}[]
 \centering
 \includegraphics[width=0.48\textwidth]{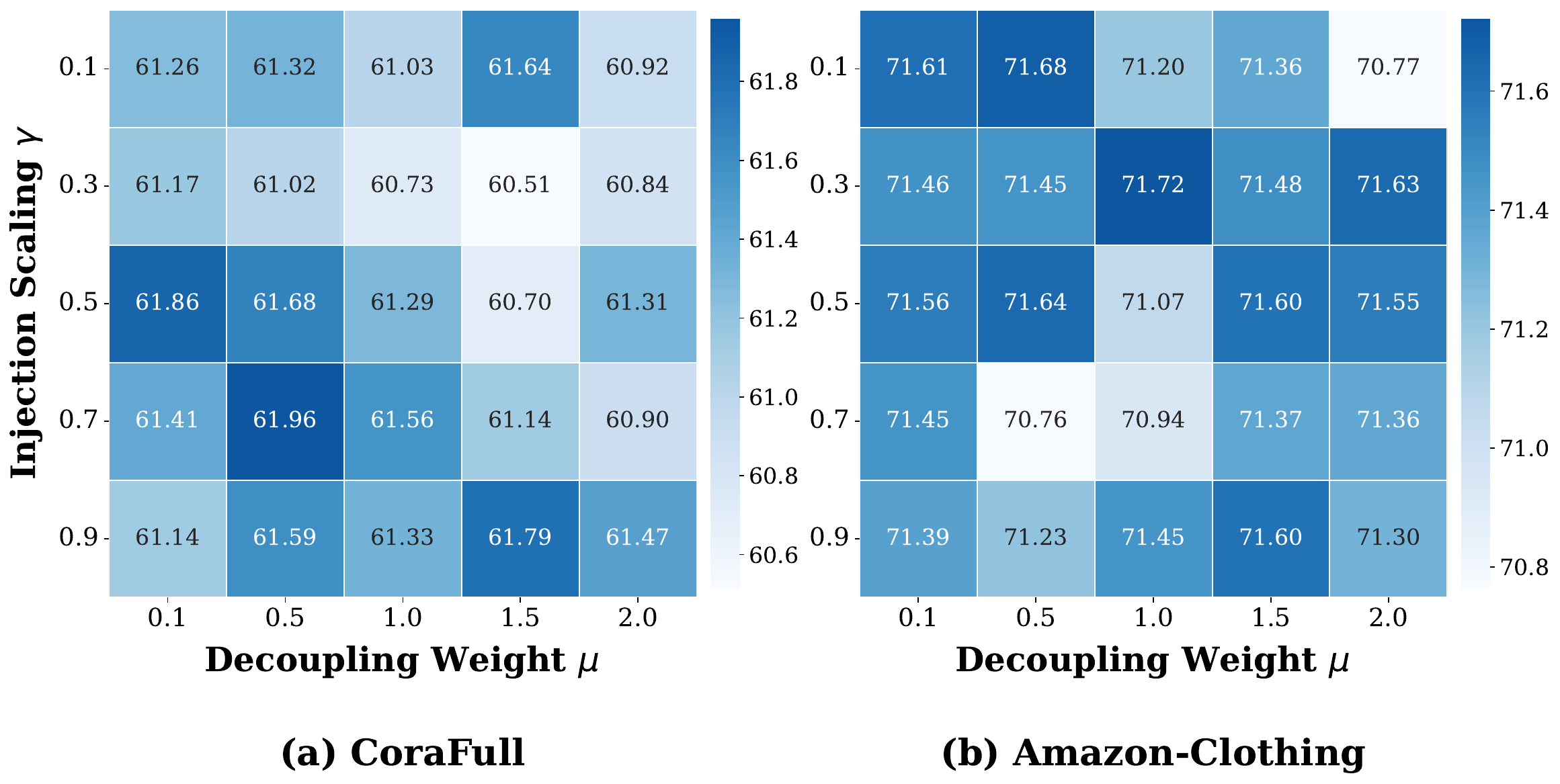}%
 \caption{Hyperparameter sensitivity analysis of FedEPD on two datasets. The heatmaps display the Macro-F1 scores across different combinations of the logit adjustment intensity $\mu$ and the injection scaling factor $\gamma$. The relatively uniform distribution of high scores demonstrates the framework's exceptional robustness to hyperparameter variations.}
 \label{hyperparameter}
 \vspace{-10pt}
\end{figure}
\subsection{Hyperparameter Analysis}
    To answer \textbf{Q4}, we investigate the sensitivity of two hyperparameters: the scaling intensity for class-aware logit adjustment ($\mu$), and the intensity parameter for injecting the global prototype into the low-frequency component ($\gamma$). 
    Figure~\ref{hyperparameter} illustrates the performance variations across different pairings of these hyperparameters on the CoraFull and Amazon-Clothing datasets. 
    Overall, FedEPD exhibits low sensitivity to these parameters, demonstrating the robustness of the dual decoupling paradigm. Specifically, $\mu$ controls the intensity of the semantic calibration for tail categories, while $\gamma$ regulates the proportion of the robust global prototype incorporated to compensate for structural deficiency. 
    The overall performance variation remains remarkably small (e.g., ranging from 60.51\% to 61.96\% on CoraFull, and 70.76\% to 71.72\% on Amazon-Clothing), providing flexibility for practical deployment. 
    Due to diverse topological properties across datasets, the absolute peak performance does not correspond to a single fixed combination. 
    However, rather than presenting a narrow optimal window, the results reflect a broad optimal plateau. Any combination within our rigorously tested grid ($\mu \in [0.1, 2.0]$ and $\gamma \in [0.1, 0.9]$) yields stable results. For practical deployment, users can flexibly set these values without extensive fine-tuning and still maintain robust classification capability.

\begin{figure*}[t]
 \centering
 \includegraphics[width=1.0\textwidth]{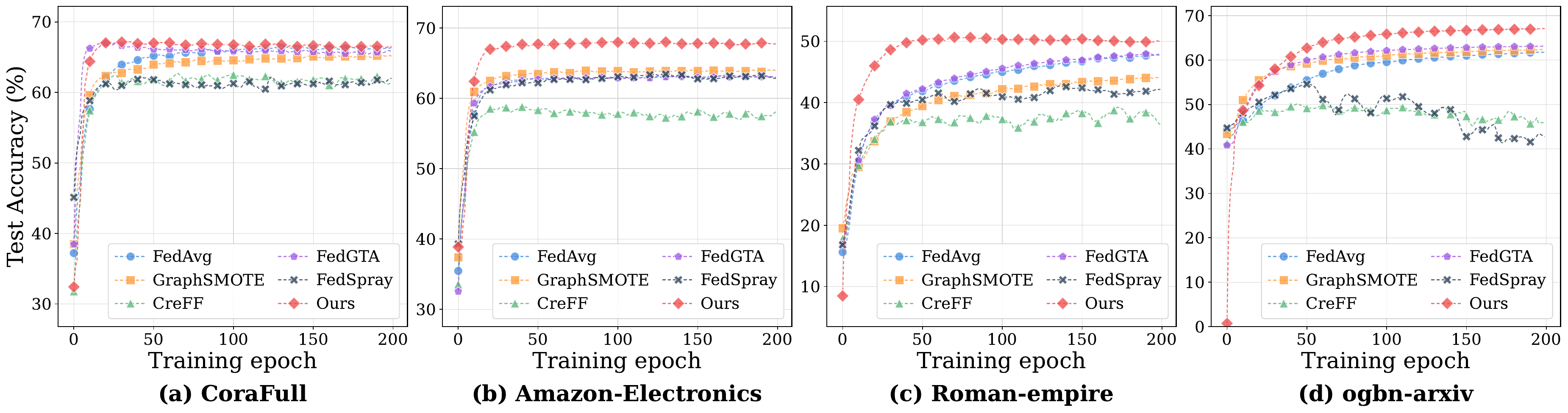}%
 \caption{Test accuracy convergence curves of our FedEPD and representative baselines over 200 training rounds on four datasets.}
 \label{curves}
 \vspace{5pt}
\end{figure*}

\begin{figure}[]
 \centering
 \includegraphics[width=0.48\textwidth]{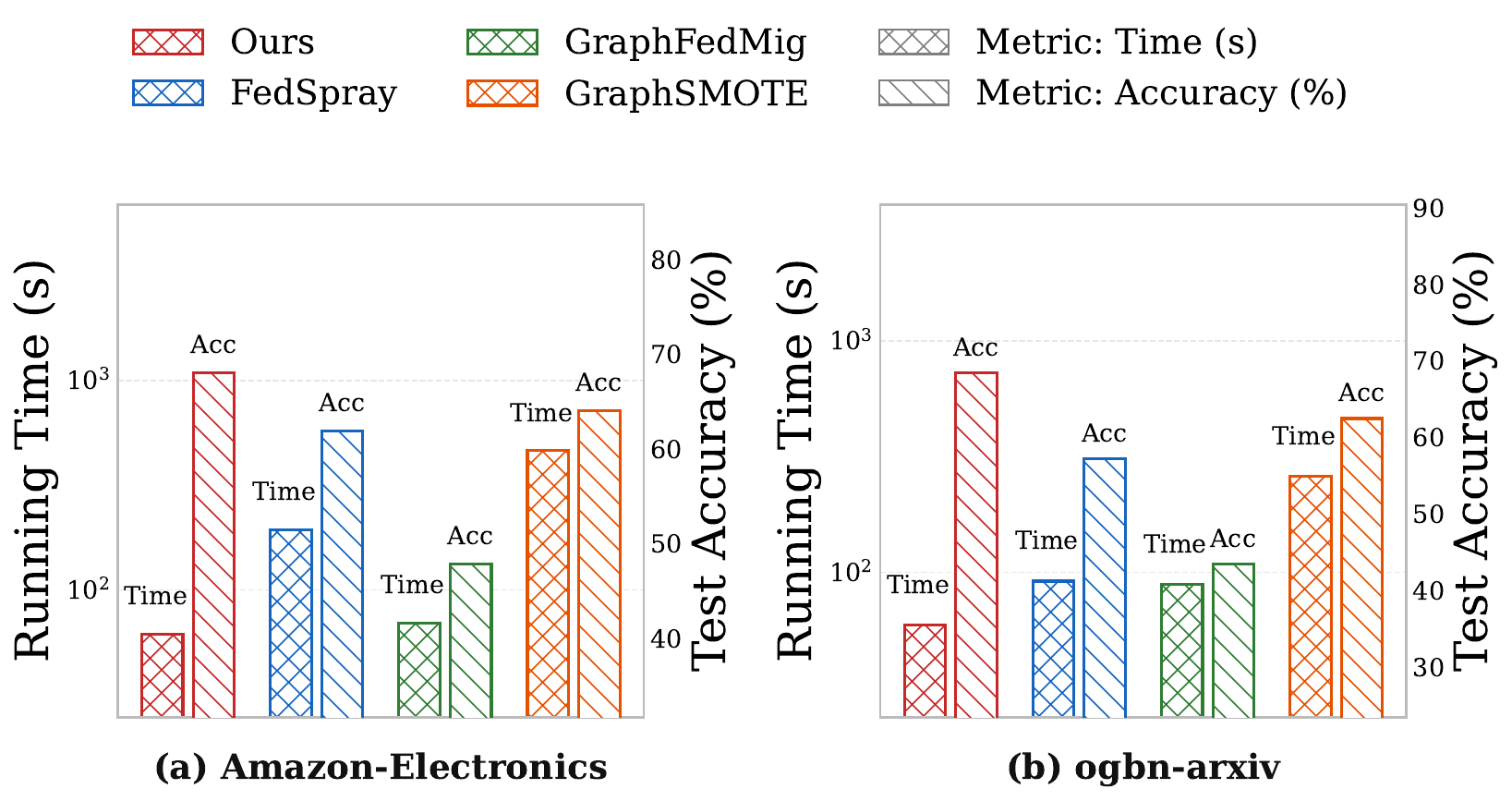}%
 \caption{Performance comparison on the (a) Amazon-Electronics and (b) ogbn-arxiv datasets. The left vertical axis indicates the running time, and the right vertical axis represents the test accuracy.}
 \label{runtime}
\end{figure}

\subsection{Efficiency Analysis}
    To answer \textbf{Q5}, we evaluate the system efficiency and convergence behavior of the proposed framework. To assess the efficiency-accuracy trade-off, we compare the end-to-end running time against the final test accuracy on the Amazon-Electronics and ogbn-arxiv datasets, as visualized in Fig.~\ref{runtime}. 
    The dual-axis bar charts reveal an inherent efficiency-accuracy trade-off among existing baselines. Lightweight methods require minimal execution time but yield lower test accuracy. Conversely, methods employing complex synthetic mechanisms to address long-tailed distributions incur substantial computational overhead, requiring significantly longer running times while resulting in suboptimal accuracy. 
    In contrast, FedEPD provides an optimized balance for this trade-off. Across both datasets, our method achieves the highest test accuracy while maintaining a running time comparable to, or lower than, the baselines with the lowest computational costs. 

    This efficiency advantage is supported by our complexity analysis. Computationally, as established in Section \ref{sec:purification}, the multi-metric topological purification requires a one-time preprocessing cost bounded by $\mathcal{O}(|\mathcal{E}_k|d + |\mathcal{E}_k| \log |\mathcal{E}_k|)$. Additionally, the iterative PPR approximation restricts the local prototype extraction overhead to $\mathcal{O}(L|\mathcal{E}_k|)$. Because these operations are executed and cached solely during the initial communication round, the subsequent local alternating optimization stages introduce negligible computational burden. Furthermore, the communication payload per round only involves transmitting the local prototype sums $\mathbf{u}_k^c \in \mathbb{R}^d$ and elite counts $m_k^c$ alongside standard model weights. This additional overhead is bounded by $\mathcal{O}(C \times d)$, which is negligible compared to the neural network parameter size $\mathcal{O}(|\Theta|)$, thereby avoiding communication bottlenecks.

    Beyond computational efficiency, we evaluate the convergence stability of the framework. As illustrated in Fig.~\ref{curves}, which visualizes the test accuracy trajectories over 200 communication rounds across four datasets, FedEPD reaches a higher accuracy plateau across evaluated scenarios, encompassing both homophilic and heterophilic graphs. 
    Specifically, while standard federated methods like FedAvg and FedGTA exhibit stable convergence plateaus, approaches addressing the long-tailed issue without dual decoupling frequently exhibit convergence fluctuations. Conversely, FedEPD maintains a stable convergence trajectory. This empirical evidence supports the efficacy of the two-stage alternating optimization. By executing the topological purification prior to the semantic calibration and protecting the representation backbone with frozen graph encoder operations, the framework isolates the gradient conflicts between majority and minority classes that degrade long-tailed learning.
    
\section{Conclusion}
\label{Conclusion}
    In this paper, we address the critical challenge of long-tailed classification in Federated Graph Learning (FGL), where structural deficiency and heterophilic noise disproportionately degrade minority-class representations during neighborhood aggregation. 
    To resolve this, we propose FedEPD, a framework operating under a dual decoupling paradigm that strictly isolates topological purification from semantic recalibration. Specifically, FedEPD executes a tuning-free, data-driven pruning mechanism that fuses local Dirichlet energy and semantic similarity via non-parametric cumulative distributions, effectively filtering spatial heterophilic edges without requiring dataset-specific thresholds. Subsequently, a server-assisted local consensus extracts structurally robust global prototypes. These prototypes are conditionally incorporated into the low-frequency components of labeled training nodes, preventing label leakage while preserving high-frequency individual node features. Coupled with a topology-aware logit adjustment, our two-stage alternating optimization isolates gradient conflicts between majority and minority classes during the representation learning and classifier calibration phases. Extensive empirical evaluations across multiple datasets demonstrate that FedEPD achieves superior performance on both homophilic and heterophilic graphs, with particularly significant accuracy gains in tail-class recovery. By delivering self-adaptive structural purification and superior convergence stability, this work establishes a rigorous mathematical and empirical foundation for deploying decoupled FGL systems in complex, highly skewed non-IID environments.

\printcredits


\section*{Declaration of Competing Interest}
The authors declare that they have no known competing financial interests or personal relationships that could have appeared to influence the work reported in this paper.

\section*{Acknowledgements}
This work was supported by the Shenzhen Fundamental Research Program,  \\China under Grant JCYJ20230807094104009.

\section*{Data availability}
The data used in this research are publicly available graph datasets.

\bibliographystyle{cas-model2-names}
\bibliography{main}



\end{document}